\documentclass[pdflatex,sn-nature]{sn-jnl}

\usepackage{graphicx}%
\usepackage{multirow}%
\usepackage{amsmath,amssymb,amsfonts}%
\usepackage{amsthm}%
\usepackage{mathrsfs}%
\usepackage[title]{appendix}%
\usepackage{xcolor}%
\usepackage{textcomp}%
\usepackage{manyfoot}%
\usepackage{booktabs}%
\usepackage{algorithm}%
\usepackage{algorithmicx}%
\usepackage{algpseudocode}%
\usepackage{listings}%
\usepackage{caption}
\usepackage{mathtools}
\usepackage{subcaption}

\theoremstyle{thmstyleone}%
\theoremstyle{thmstyletwo}%

\theoremstyle{thmstylethree}%

\begin{document}

\title[Neural Partial Differential Equations and The Illusion of Learning]{What You See is Not What You Get: Neural Partial Differential Equations and The Illusion of Learning}


\author*[1]{\fnm{Arvind} \sur{Mohan}}\email{arvindm@lanl.gov}

\author[2]{\fnm{Ashesh} \sur{Chattopadhyay}}\email{aschatto@ucsc.edu}

\author[1]{\fnm{Jonah} \sur{Miller}}\email{jonahm@lanl.gov}

\affil*[1]{\orgdiv{Computational Physics and Methods Group}, \orgname{Los Alamos National Laboratory}, \orgaddress{\street{P.O.Box 1663}, \city{Los Alamos}, \postcode{87544}, \state{New Mexico}, \country{USA}}}

\affil[2]{\orgdiv{Department of Applied Mathematics}, \orgname{University of California, Santa Cruz}, \orgaddress{\street{1156 High St}, \city{Santa Cruz}, \postcode{95064}, \state{California}, \country{USA}}}


\abstract{Differentiable Programming for scientific machine learning (SciML) has recently seen considerable interest and success, as it directly embeds neural networks inside PDEs, often called as NeuralPDEs, derived from first principle physics. Unlike large, parameterized black-box deep neural networks, NeuralPDEs promise a targeted and efficient learning approach by only representing unknown terms and allowing the rest of the known PDE to constrain the network with known physics. Owing to these strengths, there is a widespread assumption in the community that NeuralPDEs are more trustworthy and generalizable than black box models. However, like any SciML model, differentiable programming relies predominantly on high-quality PDE simulations as “ground truth” for training. In the era of foundation models, the blanket assumption of ground truth has been made for any scientific simulation that satisfactorily captures physical phenomena. However, mathematics dictates that these are only discrete numerical \textit{approximations} of the true physics in nature. 
Therefore, we must pose the following questions: Are NeuralPDEs and differentiable programming models trained on PDE simulations indeed as physically interpretable as we think? And in cases where the NeuralPDEs can successfully extrapolate, are they doing so for the right reasons? In this work, we rigorously attempt to answer these questions, using established ideas from numerical analysis and dynamical systems theory. We use (1+1)-dimensional PDEs as our test cases: the viscous Burgers equation, and the geophysical Kortveg de Vries equations. 

Our analysis shows that NeuralPDEs learn the artifacts in the simulation training data arising from the discretized Taylor Series truncation error of the spatial derivatives. Consequently, we find that NeuralPDE models are systematically biased and their generalization capability likely results from, instead of learning physically relevant quantities, a fortuitous interplay of numerical dissipation and truncation error in the training dataset and NeuralPDE, which seldom happens in practical applications. The evidence for our hypothesis is provided with mathematical theory, numerical experiments and analysis of model Jacobians.  This bias manifests aggressively even in such relatively accessible 1-D equations, raising concerns about the veracity of differentiable programming on complex, high-dimensional, real-world PDEs, and in dataset integrity of foundation models. Further, we observe that the initial condition constrains the truncation error in initial-value problems in PDEs, thereby exerting limitations to extrapolation. Additionally, we demonstrate that an eigenanalysis of the learned network weights can indicate a priori if the model will be inaccurate for out-of-distribution testing.}





\maketitle

\section{Introduction}\label{sec1}
One of the most promising avenues of machine learning (ML) research in physical sciences is the discovery of equations and mathematical relationships from data. This fundamentally differs from the philosophy of employing ML merely as a predictive tool, which is often agnostic to the physical processes that generated the dataset, typically an ordinary or partial differential equation (ODE or PDE). Such ``equation discovery" strategies have grown in popularity because they aim to exploit the exceptional representational capacity of neural networks (NNs), without eschewing the numerous benefits of explainability and robustness enjoyed in numerical solutions of traditional ODE/PDEs that closely describe the dynamics of the problem. 



Combining the strengths of PDEs and NNs belongs to a larger class of methods, best popularized by the pioneering work of Chen~\cite{chen2018neural}. Neural ODEs interpreted the learning problem as a traditional timestepper problem, but where the time derivative is modeled as a NN, instead of discretized derivatives germane to the problem, such as those found in numerical solvers. This idea has been groundbreaking for surrogate and reduced order modeling of large systems: The NN is tuned on a higher-fidelity, expensive ground truth PDE simulation (and less commonly, from observational data), which serves as the training dataset. As shown in literature~\cite{rackauckas2020universal,kidger2022neural,shen2023differentiable,sapienza2024differentiable}, the flexibility and superior representational capacity of NNs allows it to model the dynamics of the system at a far lower computational cost than the true PDE. Rapid ensemble forecasts with ML has seen a surge of interest in a variety of disciplines, particularly in weather, climate modeling, engineering design and fluid dynamics, since PDE simulations at scale are prohibitively expensive~\cite{pachalieva2022physics,ramsundar2021differentiable}. This opens up a range of applications which require fast and cheap ensemble solutions of systems with reasonable accuracy, such as feedback control, engineering design exploration, emergency and scenario analysis, and earth sciences.

Following the footsteps of NeuralODE based methods, a larger class of NeuralPDE methods emerged, which integrate PDEs with neural networks (NN) more tightly under the umbrella of \textit{differentiable programming} or DP. DP approaches PDEs and scientific computing from the broader perspective of automatic differentiation (AD) on mathematical operations, of which NN architectures are a special case. In the DP paradigm, we can compose an arbitrary sequence of mathematical operations with \textbf{known} PDEs, mathematical functions and the \textbf{unknown} components parameterized by NN. Learning the NN is accomplished by backpropagating through not just the trainable parameters, but the entire sequence of operations, which include both the known and parameterized unknown behavior of the system. Such a hybrid learning combination accomplishes two goals: a) It ensures that we can leverage existing scientific knowledge in differential equations to satisfy basic conservation laws that NNs struggle to capture, and b) Leave the more challenging, higher-order descriptions that often elude a clean analytical description, to be learned. Strategically blending equations and NNs enables \textit{targeted learning} of only the unknown quantities improves accuracy and generalization at a much lower computational cost, by reducing the need for large, physics-agnostic, complex NN architectures that are expensive to learn and prone to overfitting. As a results, numerous efforts have been taken to write traditional physics solvers in a differentiable form such that AD and backpropagation can be accomplished by parameterizing unknown or expensive terms in the PDEs. These differentiable physics solvers have shown intriguing promise across many disciplines~\cite{gelbrecht2021neural,ramadhan2020capturing,hernandez2019differentiable,csala2024physics,chakrabarti2023full,frezat2022posteriori}, but also involve significant human effort, as it is a fundamentally intrusive process that requires partial or complete rewriting of the solver in a differentiable framework, including the software machinery to efficiently compute adjoints at scale. This is a non-trivial challenge, as shown by recent efforts in weather and climate research such as NeuralGCM~\cite{kochkov2024neural}, JAX-Fluids~\cite{bezgin2023jax} and in fusion power and control systems research such as TORAX~\cite{TORAX}, where differentiable physics are extremely attractive.

Owing to these strengths, there is widespread belief in the community that NeuralPDEs are more interpretable and trustworthy: After all, the parameterizations are intimately aware of the governing PDEs where spatial derivatives are mathematically rigorous for the physics of the problem. Still, the community has acknowledged the unique challenges of training NNs within the context of ODE and PDE solvers, especially from the perspective of long-term accuracy and stability of the PDEs and ODEs. The work of Um et al.~\cite{um2020solver} focused on the temporal error accumulation of NeuralPDEs due to discretization errors when the resolution of the training data changed. They proposed a ``solver-in-loop" strategy based on the argument that temporal variation in grid-induced error has structural similarities that can be learned if the NN is embedded inside the solver generating the training data. This allows the NN to learn gradients and dynamics to reduce this error for long-term rollouts. Their intent on isolating grid discretization error meant using the same spatial numerical method for different grid resolutions in their studies. A similar focus was presented by Brandsetter~\cite{brandstetter2022message}, Lippe~\cite{lippe2024pde}, and many others~\cite{list2024temporal,zhang2024sinenet,list2025differentiability} where several methods to decrease error accumulation over time integration were proposed. However, literature is scarce in rigorously assessing the impact of numerical methods independent of the grid resolution, and their impact on the learned dynamics. 



Therefore, this work investigates the intrinsic errors and biases arising from the numerical discretization of spatial derivatives and their impact on differentiable models. We employ established protocols from computational physics and numerical analysis to verify \textit{if parameterized PDEs are indeed predicting accurate solutions for the right reasons}. We specifically focus on \textit{non-intrusive} models  where the NeuralPDE is intended to be a surrogate model of the ground truth equation, as this is often a realistic use case in many applications. Ground truth PDEs are often expensive and slow, and surrogate PDEs have long been desired for applications in ensemble forecasting, control, and rapid design iteration, where forecast speed and computational efficiency are paramount. In these cases, researchers formulate and discretize PDE terms to the best of their knowledge and parameterize the unknown terms with NN. Despite the focus on non-intrusive models, many results of this study are readily extendable to \textit{intrusive} differentiable programming models, where NN are embedded inside the ground truth PDE solver. Our key findings are as follows:

\begin{itemize}
    \item Differentiable Programming models of PDEs have a significant sensitivity to the interaction of truncation errors originating from the Taylor series expansion of numerical scheme representing the spatial derivative and the initial condition of the PDE trajectory used in training. We show this manifests even when the grid size is constant, and we \textit{can analytically represent the structure }of this error.
    \item In many NeuralPDE models that have extrapolation accuracy, we demonstrate that this is likely possible only under specific conditions where the numerical dissipation from Taylor series (TS) \textit{truncation error in the NeuralPDE matches the truncation error in the training dataset}, which happens when the same numerical scheme is used in both instances.
    \item We propose that in large training datasets, such as those used for foundation models, intrinsic sample differences in truncation error may act as an ``adversarial attack" on the model. This happens when truncation error dynamics are unique to each solver that generated them, but are not apparent as the trajectories are qualitatively and quantitatively similar. 
\end{itemize}

    

    

\section{The (Mathematical) Case for Skepticism} \label{skepticmath}
Any smooth solution\footnote{We note that many, if not most, solutions of interest are \textit{not} smooth. Nevertheless the spirit of the following discussion holds.} of a well-posed initial value problem may be thought of as a vector in an infinite-dimensional Sobolev or Hilbert space \cite{brezis2010functional}. More concretely, it may (for example) be expressed as an infinite sum over linearly independent basis functions $\Phi_i(x)$
\begin{equation}
    \label{eq:def:sum}
    f(t,x) = \sum_{i=0}^\infty c_i(t) \Phi_i(x),
\end{equation}
where the coefficients $c_i$ are given by the \textit{projection} of $f$ onto $\Phi$ via the integral
\begin{equation}
    \label{eq:projection}
    c_i(t) = \left\langle f, \Phi\right\rangle = \int_\Omega dx \Phi_i(x) f(t, x),
\end{equation}
where here we have assumed without loss of generality that the basis functions $\Phi$ are normalized such that
$$\left\langle \Phi_i, \Phi_j\right\rangle = \delta_{ij}.$$
This is the infinite-dimensional generalization of standard linear algebra and the standard vector inner product. The Fourier series is a classic example of this infinite series representation. However, any set of ``basis functions'' will do, so long as they are linearly independent (though here we assumed orthonormality), and the Fourier basis is only one such special case.

When a PDE and its solution are discretized on a computer, this may be interpreted as a truncating sum (see Eq.~\eqref{eq:def:sum}) and projecting the solution $f$ onto a finite-dimensional subspace of the infinite-dimensional solution space. This operation implies that, although we often train our neural networks assuming a numerical solution is ground truth, our training data is \textit{not} ground truth. Rather, it already carries with it this projection error, representing the finite-dimensional nature of the a given discretization of the infinite dimensional solution. 

\noindent Most NeuralPDEs in applications are trained on dataset comprised of numerical simulations, where partial derivatives are approximated by discretized numerical methods. These tend to be derived from finite volume (FV) and finite difference (FD) methods. A key aspect of these methods is using TS expansions to represent a derivative in algebraic form.
This is, of course, Eq.~\eqref{eq:projection} in the special case of the monomial basis. We make a choice on the desired accuracy of the derivative by truncating the infinite TS expansion to order $n$, and the resulting numerical method is said to be of $n^{th}$ order. Therefore, while the PDE solved in a simulation consists of derivatives that explain real-world physics, in practice, these derivatives carry an approximation error associated with TS truncation. This error is entirely a mathematical artifact of solving for physics on computers. Employing careful analysis, our paper will show how this benign error has serious consequences for the accuracy and reliability of Neural Differential Equations for predictive modeling.

If we account for truncation error, the \textit{correct} representation of a $2^{nd}$ order central difference of the second derivative of a variable $\phi$ is,
\begin{equation}
     \frac{\phi_{i-2} - 2\phi_{i-1} + \phi_{i}}{h^2} = \frac{\partial^2 \phi}{\partial x^2} \bigg|_{true} -  \sum_{i=2}^{\infty} \frac{\delta x^i}{i!} \frac{\partial^i \phi}{\partial x^i}
\end{equation}

In general, any $n^{th}$ order FD/FV representation of a derivative $p$ of $\phi$ can be written in general form,

\begin{equation}
     \mathrm{FDM(\phi)} = \frac{\partial^p \phi}{\partial x^p} \bigg|_{true} - \sum_{i=n}^{\infty} \frac{\delta x^i}{i!} \frac{\partial^i \phi}{\partial x^i}  
     \label{eqn:fdmgeneralform}
\end{equation}

The representation of numerical derivatives in Eq.~(\ref{eqn:fdmgeneralform}) indicates two key factors affecting the training data's truncation error represented by the summation term: a) The order of approximation $n$, as it affects the truncation fidelity, and b) In case of initial value problems (IVPs) that dominate PDEs, the initial condition (IC) $\phi_0$ for each simulation in the training dataset, as it affects the infinite series of the derivatives. To simplify the analysis, we will assume that the grid resolution $\delta x$ is kept constant across simulations.

\subsection*{PDE Theory}\label{sec:PDE:theory}

A differential equation with initial and boundary conditions is called a \textit{well-posed initial value problem} if there exists a unique solution, such that the solution varies continuously with initial conditions (in the variational sense) \cite{Hadamard}. Practically, well-posedness guarantees that small uncertainty in initial conditions, for a physical problem, translates to small uncertainty in solution.

The viscous Burgers' equation may be shown to form a well-posed initial value problem via the existence of an energy norm \cite{Aubin,Lions,Limaco}
\begin{equation}
    \label{eq:def:burgers:energy}
    E(t) = \int |\phi(t,x)|^2 e^{x^2/4}dx \geq 0
\end{equation}
such that 
\begin{equation}
    \label{eq:non-increasing}
    \frac{d E}{dt} \leq 0,
\end{equation}
i.e., the norm must be positive definite and non-increasing \cite{Hadamard}. Alternatively, a maximum principle may be applied \cite{HopfLemma}. Similar arguments may be made for the geophysical KdV equation~\cite{geyer2018shallow} (though the energy norm looks different) and results are summarized for equations of the standard KdV form in \cite{Akhunov_2019}. If a neural PDE augments either equation, and learns the numerical viscosity inherent to a numerical solution in training data, condition \eqref{eq:non-increasing} may be violated and the neural realization may become ill-posed (and thus no longer predictive). 

At the level of a linearization or discretization of a given PDE, von Neumann stability analysis may be used to examine stability, which represents the final condition in well-posedness, the continuity of solutions with respect to initial data \cite{crank1947practical}. Our analysis of the eigenvalues of the Jacobian of the neural network is a realization of this form of analysis. If errors compund with time, then the method is not linearly stable. We note that linear stability does not guarantee nonlinear stability.

\subsection{Burgers Equation} \label{skepticmath:burgers}

Consider the 1-D viscous Burgers equation as a candidate problem for analysis. In many works, including this one, the Burgers equation is used as a toy model due to its relative simplicity. When viscosity is weak, it is the simplest hyperbolic nonlinear equation, exhibiting varying wave speeds and shock structure. The challenges seen in such simple cases directly extend to more complex equations, which drive many applications. The most common use case for NeuralPDEs is parameterizing missing/unknown physics in a partially known PDE with a $NN$. The variable of interest $\phi$ is the velocity, which evolves per Eq.~(\ref{eqn:trueEqn}). We propose a contrived case where the diffusion term is the ``unknown" that needs to be learned by a Neural PDE, to obtain predicted velocity $\hat{\phi}$. We show this in Eq.~(\ref{eqn:learnedEqn}) where the NeuralPDE retains the time derivative and advection terms, but the diffusion term is parameterized by a $NN$ that takes $\hat{\phi}$ as an input. The goal of this $NN$ is to modulate $\hat{\phi}$ such that the diffusive behavior seen in the ground truth is adequately modeled to obtain the correct velocity predictions, such that $\hat{\phi} \rightarrow \phi$ when training converges.

\begin{align}
        \frac{\partial \phi}{\partial t} &= \phi \frac{\partial \phi}{\partial x} + \nu \frac{\partial^{2} \phi}{\partial x^{2}} \label{eqn:trueEqn} \\ 
        \frac{\partial \hat{\phi}}{\partial t} &= \hat{\phi} \frac{\partial \hat{\phi}}{\partial x} +  f_{NN}(\hat{\phi}) \label{eqn:learnedEqn}
\end{align}

Literature has extensively employed many versions of this problem structure with several successes, notably Refs~\cite{gelbrecht2021neural,rackauckas2020universal}. Yet, \textit{an implicit assumption made is that the terms to be learned from the ground truth represent only the physics}. In contrast, mathematics dictates that these are only \textit{approximations} of the true physics, as a function of $n$, $\delta x$, and $\phi$.
Accounting for this mathematical reality, the true learning problem can be written as,

\begin{align}
    \frac{\partial \phi}{\partial t} &= \phi \frac{\partial \phi}{\partial x} \bigg|_{FDM} + 
    \sum_{i=p}^{\infty} \frac{\delta x^i}{i!} \frac{\partial^i \phi}{\partial x^i}  
    +  \nu \frac{\partial^{2} \phi}{\partial x^{2}} \bigg|_{FDM} + \sum_{i=q}^{\infty} \frac{\delta x^i}{i!} \frac{\partial^i \phi}{\partial x^i}    \label{eqn:FD_trueEqn} \\ 
    \frac{\partial \hat{\phi}}{\partial t} &= \hat{\phi} \frac{\partial \hat{\phi}}{\partial x} \bigg|_{FDM} + \sum_{i=k}^{\infty} \frac{\delta x^i}{i!} \frac{\partial^i \hat{\phi}}{\partial x^i} + f_{NN}(\hat{\phi}) \label{eqn:FD_learnedEqn}
\end{align}

All terms are discretized with finite differences of various orders, indicated by the superscripts $p$, $q$, and $k$ respectively.
These equations offer an accurate representation of the learning problem, where the true form of the discrepancy $\varepsilon$ between the training data and NeuralPDE training, which must be minimized, can be written as:
\begin{equation}
\begin{aligned}
    \varepsilon &= \underbrace{\phi \frac{\partial \phi}{\partial x} \bigg|_{FDM} -  \hat{\phi} \frac{\partial \hat{\phi}}{\partial x} \bigg|_{FDM}}_{\text{Advective error}}  +  \underbrace{\nu \frac{\partial^{2} \phi}{\partial x^{2}} \bigg|_{FDM} - f_{NN}(\hat{\phi})}_{\text{Diffusive error}} + 
    \\ 
    &\underbrace{\sum_{i=p}^{\infty} \frac{\delta x^i}{i!} \frac{\partial^i \phi}{\partial x^i} - \sum_{i=k}^{\infty} \frac{\delta x^i}{i!} \frac{\partial^i \hat{\phi}}{\partial x^i} + \sum_{i=q}^{\infty} \frac{\delta x^i}{i!} \frac{\partial^i \phi}{\partial x^i}}_{\text{Numerical error}}. \label{eqn:discrepancy_burgers_NeuralPDE}
    \end{aligned}
\end{equation}


We emphasize that Eqs.~(\ref{eqn:trueEqn}) and~(\ref{eqn:learnedEqn}) represent only our \textit{idealized expectation} of what is being learned. The unstated assumption in the literature is to think of the discrepancy $\varepsilon$ only as the first two terms in Eq.~(\ref{eqn:discrepancy_burgers_NeuralPDE}), such that $\varepsilon \rightarrow 0$ as $f_{NN}(\hat{\phi})$ approximates the diffusion term during training. Yet, the numerical error has multiple terms dictated by the training data's spatial discretization and the initial conditions used to generate them, even when grid size and time integration schemes are the same for Eqs.~(\ref{eqn:trueEqn}) and~(\ref{eqn:learnedEqn}).

\subsection{Geophysical Kortveg de Vries Equation}
To show that the effect of numerical discretization and initial conditions is agnostic to the type of derivatives, we also analyze the geophysical Kortveg de Vries (gKdV) equation. The gKdV is a variation of the classic Kortveg de Vries equation, which is a complex nonlinear system that predicts the evolution and interaction of waves in a medium. The KdV is popular due to its practical relevance, and it has an advective term and a dispersive term, which is a third-order derivative. The dispersive term has qualitatively different behavior compared to first and second-order derivatives, as it gives rise to oscillatory dynamics. In general, even-order derivatives exhibit diffusive and hyper-diffusive behavior, while odd-order derivatives exhibit varying degrees of dispersive, oscillatory behavior. An additional complexity in gKdV that does not manifest as strongly in the Burgers equation is the extremely transient nature of the solution due to interactions between multiple wavefronts. This provides an interesting test case to study how the effect of learning truncation error manifests in long-time rollout stability and accuracy of the solutions with varying IC and numerical discretization.

The gKdV was developed in Geyer and Quirchmayr~\cite{geyer2018shallow} by modifying the kdV to account for the dynamics of equatorial tsunamis in the ocean. A key difference in the gKdV is the addition of another advective term with a coriolis force $\omega_0$, that is obtained from observational data. The gKdV can thus be written as
\begin{equation}
    \frac{\partial \phi}{\partial t} = \frac{\partial \phi}{\partial x} \left(\omega_{0} - \frac{3}{2} \phi \right) - \frac{1}{6} \frac{\partial^{3} \phi}{\partial x^{3}}
    \label{eqn:gkdvtrueEqn}
\end{equation}

The quantity learned in this equation is the first term since the $\omega_{0}$ is based on observational data~\cite{geyer2018shallow}. This acts as an excellent demonstration problem for realistic applications in earth sciences, where parameterizing PDEs from data is a key aspect of model construction. The NeuralPDE can be written as,

\begin{equation}
    \frac{\partial \phi}{\partial t} = f_{NN}(\hat{\phi}) - \frac{1}{6} \frac{\partial^{3} \phi}{\partial x^{3}}
    \label{eqn:gkdvlearnedEqn}
\end{equation}

Where the $f_{NN}(\cdot)$ is a neural network that accepts $\hat{\phi}$ as an input to make predictions for a term that satisfies the coriolis correction in the data. Similar to the Burgers equation  Eqs.~(\ref{eqn:FD_learnedEqn}) - (\ref{eqn:discrepancy_burgers_NeuralPDE}) we can write an explicit form for the gKdV in Finite difference.

\begin{align}
    \frac{\partial \phi}{\partial t} &= \left(\omega_{0} - \frac{3}{2} \phi \right) \frac{\partial \phi}{\partial x} \bigg|_{FDM} + 
    \sum_{i=p}^{\infty} \frac{\delta x^i}{i!} \frac{\partial^i \phi}{\partial x^i}  
    -  \frac{1}{6} \frac{\partial^{3} \phi}{\partial x^{3}} \bigg|_{FDM} + \sum_{i=k}^{\infty} \frac{\delta x^i}{i!} \frac{\partial^i \phi}{\partial x^i}    \label{eqn:FD_trueEqn}. \\ 
    \frac{\partial \hat{\phi}}{\partial t} &= f_{NN}(\hat{\phi}) -  \frac{1}{6} \frac{\partial^{3} \phi}{\partial x^{3}} \bigg|_{FDM} + \sum_{i=q}^{\infty} \frac{\delta x^i}{i!} \frac{\partial^i \phi}{\partial x^i}
     \label{eqn:FD_gKdV_learnedEqn}.
\end{align}

In this case, the discrepancy that $f_{NN}$ minimizes is given by:

\begin{equation}
\begin{aligned}
    \varepsilon &= \underbrace{\left(\omega_{0} - \frac{3}{2} \phi \right) \frac{\partial \phi}{\partial x} \bigg|_{FDM} -  f_{NN}(\hat{\phi})}_{\text{Advective error}}  
    -  \underbrace{ \frac{1}{6} \frac{\partial^{3} \phi}{\partial x^{3}} \bigg|_{FDM}  + \frac{1}{6} \frac{\partial^{3} \hat{\phi}}{\partial x^{3}} \bigg|_{FDM}  }_{\text{Dispersive error}} \\
    &\quad + \underbrace{\sum_{i=p}^{\infty} \frac{\delta x^i}{i!} \frac{\partial^i \phi}{\partial x^i} + \sum_{i=k}^{\infty} \frac{\delta x^i}{i!} \frac{\partial^i \hat{\phi}}{\partial x^i} - \sum_{i=q}^{\infty} \frac{\delta x^i}{i!} \frac{\partial^i \phi}{\partial x^i}}_{\text{Numerical error}}.
    \label{eqn:discrepancy_gKdV_NeuralPDE}
\end{aligned}
\end{equation}

The interaction between the truncated terms is qualitatively different from the Burgers case as we parameterize the advective term, and retain the dispersive term as the unknown. Furthermore, we train these NeuralPDE models differently (autoregressive vs one-shot) to demonstrate that the observations made in this work are not merely an artifact of one problem, derivative, or training methodology and are instead structural errors in PDE simulation data, as dictated by the derivation.
We will now present several numerical experiments that show the outsized impact these errors have on model generalization and stability. In both these equations, since the truncated terms consist of both $\frac{\delta x^i}{i!}$ (a consequence of the choice of numerical method) and $\frac{\partial^i \phi}{\partial x^i}$ (a consequence of initial condition $\phi_0$), we present a series of experiments that will demonstrate the impact of each of these factors in isolation. This may have nontrivial implications for the interpretability of differentiable programming for learning PDEs and PDE foundation models~\cite{ye2024pdeformer,herde2024poseidon} trained on simulation data from disparate sources. The details of training dataset generation and the ML models are presented in the Appendix~\ref{app:methods}. We also ensured that none of the numerical scheme combinations used in both the NeuralPDEs and ground truth suffered from intrinsic numerical instabilities. Therefore, any discrepancies observed are solely due to the NeuralPDE training dynamics. 

\section{Results}\label{sec:results}

\subsection{Burgers Equation:}\label{sec:results:burgers}

We demonstrate results on our first test case, the Burgers equation problem described in Eqs.~(\ref{eqn:trueEqn}) and (\ref{eqn:learnedEqn}). The Burgers equation contains first- and second-order derivatives representing advective and diffusive properties. The NeuralPDE in this problem is trained autoregressively to make predictions one timestep $\delta t$ at a time, using only the previous timestep as the input. i.e. $\phi_{t} = \mathrm{NeuralPDE}(\phi_{t-1})$. To predict for the full rollout time T,  the trained NeuralPDE takes the IC, $\phi_{0}$ as an input, and is rolled out iteratively for $n$ timesteps such that $n = T / \delta t$. This strategy is common in many scientific ML models for PDEs, particularly climate models~\cite{kochkov2024neural,chattopadhyay2023long,guan2024lucie,watt2024ace2,lam2022graphcast,bi2023accurate,pathak2022fourcastnet} and additional details about model training are provided in the appendix. We show two experiments with varying numerical schemes in the advective and diffusive term of both the ground truth and the NeuralPDE.

\begin{enumerate}
\item \textbf{True PDE} \textit{Advective}: 1st order backward, \textit{Diffusive}: 6th order central, 

\noindent \textbf{NeuralPDE} \textit{Advective}:  1st order backward  
\item \textbf{True PDE} \textit{Advective}: 1st order backward, \textit{Diffusive}: 6th order central

\noindent \textbf{NeuralPDE} \textit{Advective}:  6th order central 
\end{enumerate}

Referring to the discussion in Ref.~\ref{skepticmath:burgers}, Expt 1 is the case where $p = k$ in Eq.~(\ref{eqn:FD_learnedEqn}). This is because the dynamics of the truncated terms are exactly the same for the advective terms in the true and NeuralPDE, when the same numerical method is employed. As a result, the numerical error in Eq.~(\ref{eqn:discrepancy_burgers_NeuralPDE}) for discrepancy $\varepsilon$ is reduced to $\sum_{i=q}^{\infty} \frac{\delta x^i}{i!} \frac{\partial^i \phi}{\partial x^i}$. In contrast, Expt 2 employs a $6th$ order central numerical scheme in the NeuralPDE advective term, such that $p \neq k$ and hence the $\varepsilon$ numerical error retains its full form as shown in  Eqn.~\ref{eqn:discrepancy_burgers_NeuralPDE}. 

\begin{figure}
     \centering
         \includegraphics[width=0.75\textwidth]{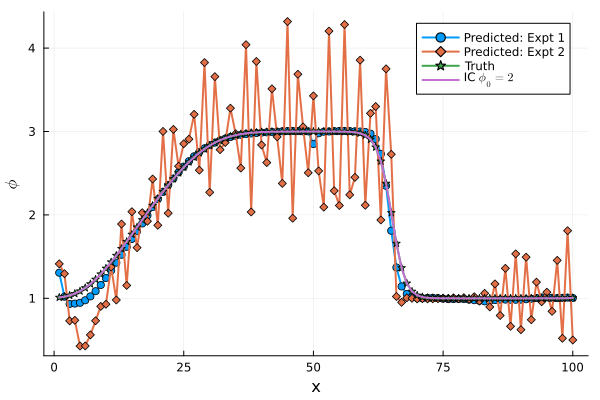}
         \caption{Ground Truth Expts 1 and 2 for training $\phi_{0} = 2$}
         \label{fig:pred_burgers_autoreg_train}
\end{figure}

\begin{figure}
     \centering
     \begin{subfigure}[b]{0.75\textwidth}
         \centering
         \includegraphics[width=\textwidth]{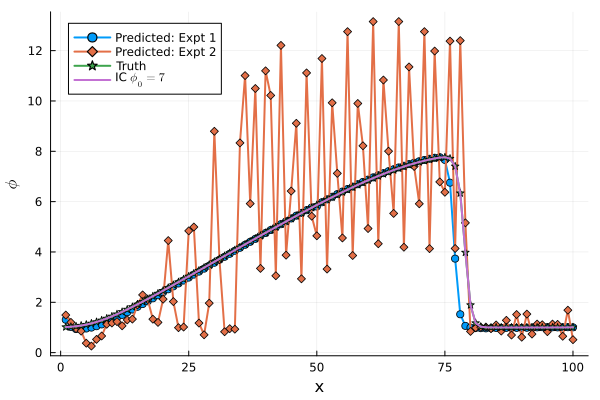}
         \caption{NeuralPDE predictions Expts 1 and 2 for test $max(\phi_{0}) = 7$}
         \label{fig:pred_burgers_autoreg_u0_7}
     \end{subfigure}
     \hfill
     \begin{subfigure}[b]{0.75\textwidth}
         \centering
         \includegraphics[width=\textwidth]{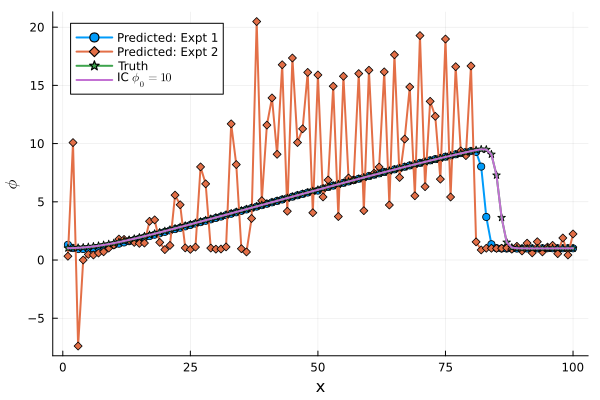}
         \caption{NeuralPDE predictions Expts 1 and 2 for test $max(\phi_{0}) = 10$}
         \label{fig:pred_burgers_autoreg_u0_10}
     \end{subfigure}
     \hfill
     \begin{subfigure}[b]{0.75\textwidth}
         \centering
         \includegraphics[width=\textwidth]{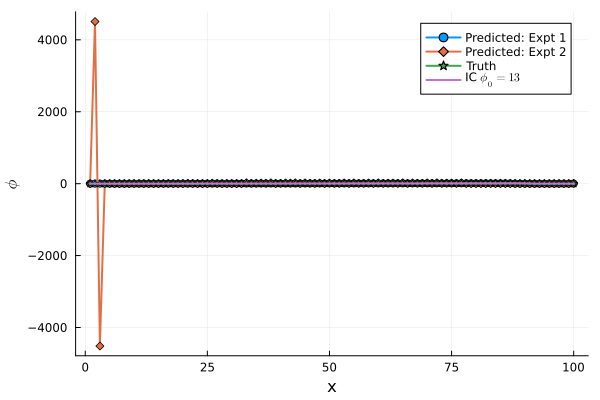}
         \caption{NeuralPDE predictions Expts 1 and 2 for test $max(\phi_{0}) = 13$}
         \label{fig:pred_burgers_autoreg_u0_13}
     \end{subfigure}     
        \caption{Burgers NeuralPDE model sensitivity when $p=k$ (Expt 1) and $p \neq k$ (Expt 2) for training IC and unseen IC. Expt 2 shows degraded performance even for training IC, and collapses at some ICs, while Expt 1 remains stable and relatively accurate.}
        \label{fig:pred_burgers_autoreg_test}
\end{figure}

\begin{figure}
    \centering
    \includegraphics[width=0.75\linewidth]{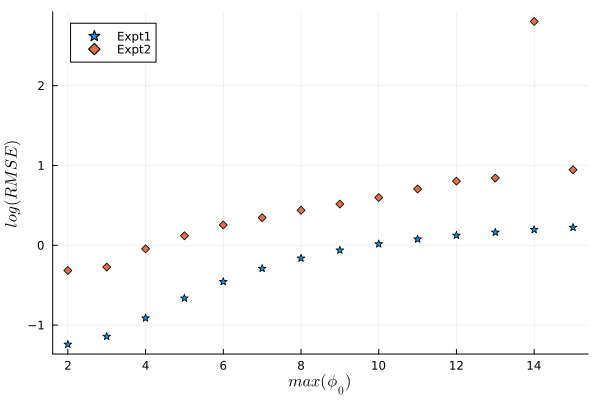}
    \caption{Log of RMS error growth with variance in IC for Expt 1 and 2. Expt 2 ($p \neq k$) shows aggressive increase in error compared to Expt 1 ($p=k$).}
    \label{fig:errorgrowth_burgers_autoreg}
\end{figure}

The NeuralPDE models are trained on a single trajectory generated by an IC $\phi_{0}^{train}$, which is kept constant for all experiments. All hyperparameters, training methods and model architectures between experiments are identical for consistency in comparisons, with the only difference being the numerical methods as described above. First, we present the predictions of the trained NeuralPDE models on $\phi_{0}^{train}$ in Fig.~\ref{fig:pred_burgers_autoreg_train} and varying $\phi_{0}^{test}$ in Fig.~\ref{fig:pred_burgers_autoreg_test}. All plots shown in this section are for the final state of the system at $t=T$.
In the case of Expt 1, its tempting to attribute the discrepancy of the location of the discontinuity to ``generalization error" of $f_{NN}$, as is often the tendency in deep learning. Instead, we posit that this a structural error arising from $\varepsilon$ in Eq.~(\ref{eqn:discrepancy_burgers_NeuralPDE}). Even though $p=k$ cancels out the truncation errors of the advective terms of the true PDE and NeuralPDE, the error from discretizing the diffusive term in the true PDE still manifests. The $\varepsilon$ is now dominated only by the term $\sum_{i=q}^{\infty} \frac{\delta x^i}{i!} \frac{\partial^i \phi}{\partial x^i} $ associated with  $\frac{\partial^{2} \phi}{\partial x^{2}} \bigg|_{FDM}$. Since $\delta x$ is constant, this truncation term is solely a function of $\phi$, i.e., in this case, the $\phi_{0}^{train} = 2$ used in training.  As a result, the theory implies that $f_{NN}$ learns $\frac{\partial^i \phi_{0}^{train}}{\partial x^i}$ corresponding to a specific wavespeed that struggles to generalize when the quantity $ \left( \phi_{0}^{train} - \phi_{0}^{test} \right)$ grows larger. Hence, the IC used in training a NeuralPDE is also a source of discretization error, in addition to the choice of numerics in the spatial derivatives. This phenomenon is readily observed in Fig.~\ref{fig:errorgrowth_burgers_autoreg}, which shows the prediction errors for NeuralPDEs in both experiments when $\phi_{0}^{test} = 3 \rightarrow 15$. While Expt 1 has lower error growth rate than Expt 2, it still monotonically increases with $\phi_{0}^{test}$.
In the case of Expt 2, the accuracy significantly degrades. The model is unstable with an increase in $ \left( \phi_{0}^{train} - \phi_{0}^{test} \right)$ because, just as Expt 1, the $f_{NN}$ learns truncation artifacts from the IC, but also while being simultaneously subjected to a rapid error growth from the interaction of the $6^{th}$ order and $1^{st}$ order schemes when $p \neq k$. We also performed multiple versions of these experiments with different numerical methods, and obtained qualitatively similar results. Therefore, only these two experiments are shown here for brevity.


\subsection{Geophysical Kortveg de Vries Equation:}\label{sec:results:gkdv}
We now replicate the experiments for the gKdV equation, which has substantially different properties and dynamics than the Burgers equation. We train this model in a one-shot fashion as autoregressive training is too unstable to usefully train. In this approach, the NeuralPDEs takes $\phi_{0}$ and time horizon of prediction $T$ as inputs, and predict $n = T / \delta t$ timesteps in one shot. The NeuralPDE is trained such that only the prediction at final time instant $T$ is minimized, such that the $f_{NN}$ learns the entire trajectory instead of the mapping between subsequent timesteps as opposed to the autoregressive model. Since this is a qualitatively different model, we perform the same analysis from the previous section to demonstrate that the failure modes we observe there also apply to one-shot models from non-trivial PDEs

Once again, we perform two identical experiments where the NeuralPDE $f_{NN}$ learns the advection term. The only difference between the experiments is the numerical discretization of the dispersive term in the NeuralPDE. Since this problem is qualitatively much harder to predict than the Burgers equation due to non-stationarity and constructive-destructive interference between waves, we do not expect either model to retain accuracy when tested for IC far from the training regime. However, it allows us to study the impact of numerical discretization when the PDE complexity increases. All plots shown in this section are for the final state of the system at $t=T$.
\begin{enumerate}
\item \textbf{True PDE} \textit{Advective}: 2nd order central, \textit{Dispersive}: 2nd order central, 

\noindent \textbf{NeuralPDE} \textit{Dispersive}:  2nd order central 
\item \textbf{True PDE} \textit{Advective}: 2nd order central, \textit{Dispersive}: 2nd order central, 

\noindent \textbf{NeuralPDE} \textit{Dispersive}:  6th order central
\end{enumerate}

\begin{figure}
    \centering
    \includegraphics[width=0.75\linewidth]{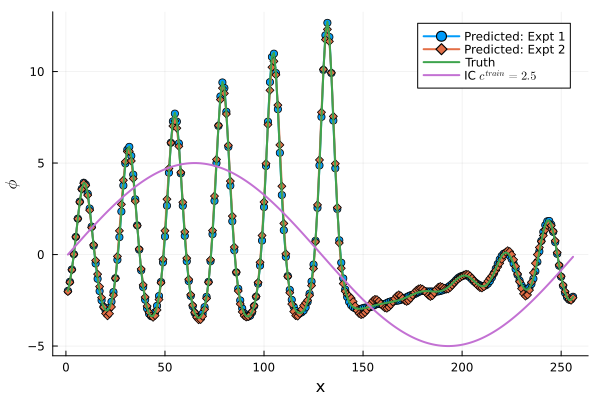}
    \caption{gKdV Expt 1 and Expt 2 Predictions for $c^{train} = 2.5$}
    \label{fig:gKdV_pred_train}
\end{figure}

\begin{figure}
     \centering
     \begin{subfigure}[b]{0.75\textwidth}
         \centering
         \includegraphics[width=\textwidth]{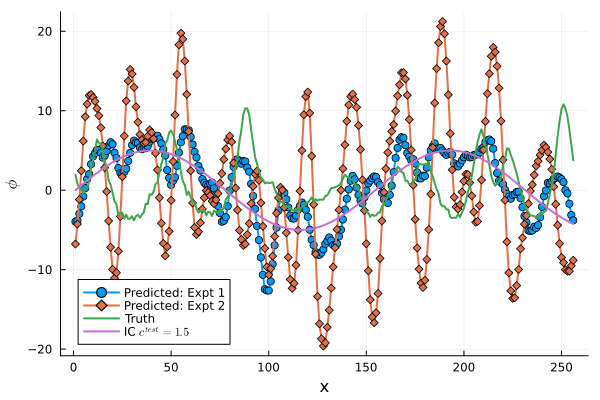}
         \caption{NeuralPDE predictions Expts 1 and 2 for test $c = 1.5$}
         \label{fig:pred_gKdV_oneshot_test_c2}
     \end{subfigure}
     \hfill
     \begin{subfigure}[b]{0.75\textwidth}
         \centering
         \includegraphics[width=\textwidth]{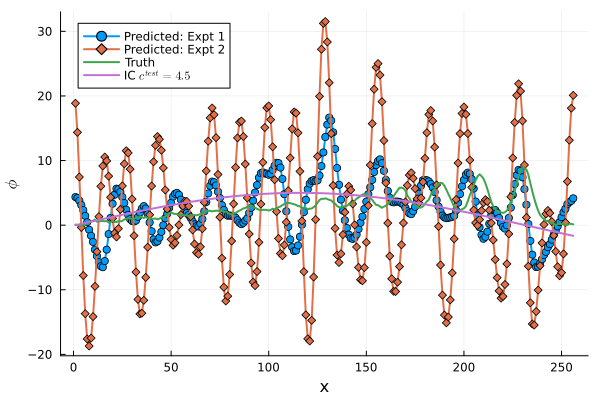}
         \caption{NeuralPDE predictions Expts 1 and 2 for test $c = 4.5$}
         \label{fig:pred_gKdV_oneshot_test_c7}
     \end{subfigure}
     \hfill
     \begin{subfigure}[b]{0.75\textwidth}
         \centering
         \includegraphics[width=\textwidth]{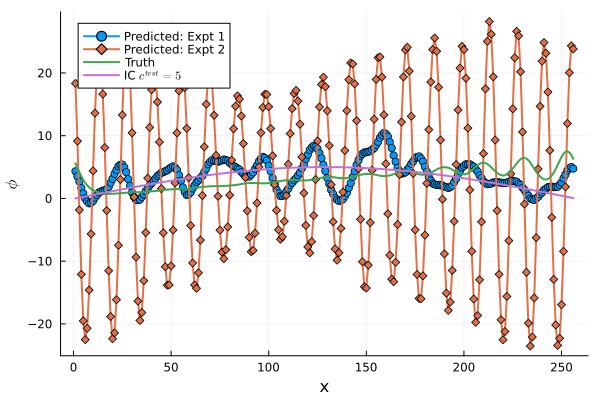}
         \caption{NeuralPDE predictions Expts 1 and 2 for test $c = 5$}
         \label{fig:pred_gKdV_oneshot_test_c8}
     \end{subfigure}     
        \caption{gKdV NeuralPDE model sensitivity when $p=k$ (Expt 1) and $p \neq k$ (Expt 2) for unseen $c$.}
        \label{fig:pred_gKdV_oneshot_test}
\end{figure}

The discrepancy $\varepsilon$ learned by $f_{NN}$ is described by Eq.~(\ref{eqn:discrepancy_gKdV_NeuralPDE}). The initial condition for gKdV is given by $u_{0} = - A \sin (x/c + \pi)$, with two free parameters $A$ and $c$. $A$ is the prescribed scalar amplitude, and $c$ is the prescribed wavespeed, on which $u_{0}$ has a nonlinear dependence. Therefore, we vary $c$ in the training and testing phase to study model generalization. Similar to Burgers, the training is performed on a single trajectory where $c^{train} = 2.5$, and testing $c^{test} = 1.5, 2, 3, 3.5, 4, 4.5, 5$. Figure~\ref{fig:gKdV_pred_train} shows the performance of models in Expt 1 and Expt 2 on the training data $c^{train} = 2.5$. Both models capture the overall trend well, but Expt 2 suffers from unphysical oscillations in $x \approx 150-200$. While this seems less concerning at this stage, it is apparent that the choice of a $6^{th}$ order central scheme in Expt 1 (as opposed to the $2^{nd}$ order central in Expt 1) has an impact on the learned model even in the training set. The impact of this choice is magnified in test conditions, which we show in Fig.~\ref{fig:pred_gKdV_oneshot_test}.

\begin{figure}
    \centering
    \includegraphics[width=0.75\linewidth]{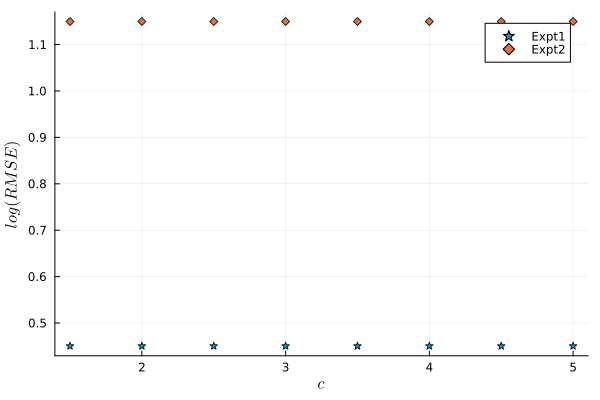}
    \caption{Log of RMS error growth with variance in $c$ for Expt 1 and 2. Expt 2 ($p \neq k$) shows consistent increase in error compared to Expt 1 ($p=k$)}
    \label{fig:errorgrowth_gKdV_oneshot}
\end{figure}

In the test regime where we show predictions for models in Expt 1 and Expt 2 for selected $c^{test} = 1.5, 4.5, 5.0$, we notice that Expt 2 has severe instabilities compared to Expt 1, which increases with $c^{test} - c^{train}$. The gKdV has an extremely nonlinear response to the initial wavespeed $c$, specifically due to the $3^{rd}$ order dispersive term that is responsible for oscillatory dynamics. The dispersive term is particularly susceptible to destabilizing the entire equation, due to its tendency of odd-powered derivatives to magnify tiny errors incurred from the training process, such as the $p \neq k$ scenario in Eq.~(\ref{eqn:discrepancy_gKdV_NeuralPDE}). In contrast, the diffusive term in Burgers was a known quantity which added some numerical ``damping" effect to the NeuralPDE, while the advective term was learned. In the gKdV, the effect of the Taylor series truncation being ``hardwired" to the training IC is also more pronounced. The model for Expt 1 which shows nearly perfect accuracy for the training conditions in Fig.~\ref{fig:gKdV_pred_train}, struggles with larger instabilities when $c^{test}$ varies, leading to unphysical oscillatory growth. Finally, we show the log of RMSE error for both experiments with changing $c^{test}$ is shown in Fig.~\ref{fig:errorgrowth_gKdV_oneshot}, where we observe a strong and consistent discrepancy between $p = k$ and $p \neq k$ scenarios.


\subsection{Eigenstability Analysis of Model Jacobians}
In this section, we take inspiration from linear stability analysis of PDEs using eigenvalue analysis.  The goal is to develop a diagnostic measure that informs us a priori that a neuralPDE may exhibit an increased error growth due to a difference in data discretization in the ground truth or due to a difference in the initial conditions. Here, we start with the general structure of the neuralPDEs used in this paper, i.e.,

\begin{align}
    \frac{\partial \hat{\phi}}{\partial t} = \mathbf{G}\left(\hat{\phi}\right)+f_{NN}\left(\hat{\phi}\right),
\end{align}

where $\mathbf{G}$ is the physics-based operator, e.g., the advection term in Burger's equation or the dispersion term in the KdV equation. In discrete form, this can be written as:
\begin{align}
    \hat{\phi}\left(t+\Delta t\right)=\underbrace{\hat{\phi}(t)+\int_{t=t}^{t=t+\Delta t}\left(\mathbf{G}\left(\hat{\phi}\right)+f_{NN}\left(\hat{\phi}\right)\right)dt}_{\mathbf{H}\left[\hat{\phi}(t)\right]},
    \label{eq:discrete_time_step}
\end{align}
where the entire right-hand-side of Eq.~(\ref{eq:discrete_time_step}) is represented by the operator, $\mathbf{H}\left[\phi(t)\right]$.

This allows us to do linear stability analysis of the discrete dynamical systems given by:
\begin{align}
    \hat{\phi}(t+\Delta t)=\mathbf{H}\left[\hat{\phi}(t)\right].
    \label{eq:discrete_dyn_with_NN}
\end{align}

We know that the initial error growth rate at $t=t_0$ is given by the largest eigenvalue of the Jacobian, $\mathbf{J}=\frac{\partial \mathbf{H}\left(\hat{\phi(t)}\right)}{\partial \hat{\phi(t)}}\bigg \rvert_{t=t_0}$. If the eigenvalue is within the unit circle, then the dynamical system given by Eq.~(\ref{eq:discrete_dyn_with_NN}) would yield a stable solution near $t=t_0$, i.e. for short-term forecasting, while if the eigenvalue is outside the unit circle then the dynamical system would go unstable. The error growth rate at $t=t_0$ is given by the magnitude of the largest eigenvalue of $\mathbf{J}$, $|\lambda_{max}|$. Generally, between two models, if the largest eigenvalue of $\mathbf{J}$ has a higher magnitude then we can expect the error to grow more quickly than the one with the lower magnitude. For the KdV equation, unlike the Burger's equation where we perform autoregressive integration in time, we perform one-shot forecasting as is done in many applications of SciML~\cite{li2020fourier,takamoto2022pdebench}. In that case, we compute a cumulative Jacobian, $\mathbf{J}_c=\frac{\partial \mathbf{H}\left(\hat{\phi}(t+100\Delta t)\right)}{\partial \hat{\phi(t)}}\bigg \rvert_{t=t_0}$, as the prediction horizon is 100 steps. While we cannot comment on the stability of the model from the largest eigenvalue of $\mathbf{J}_c$, we can still say that the error growth rate would be proportional to the largest eigenvalue of $\mathbf{J}_c$. 

\begin{figure}[h!]
    \centering
    \includegraphics[width=1.2\linewidth]{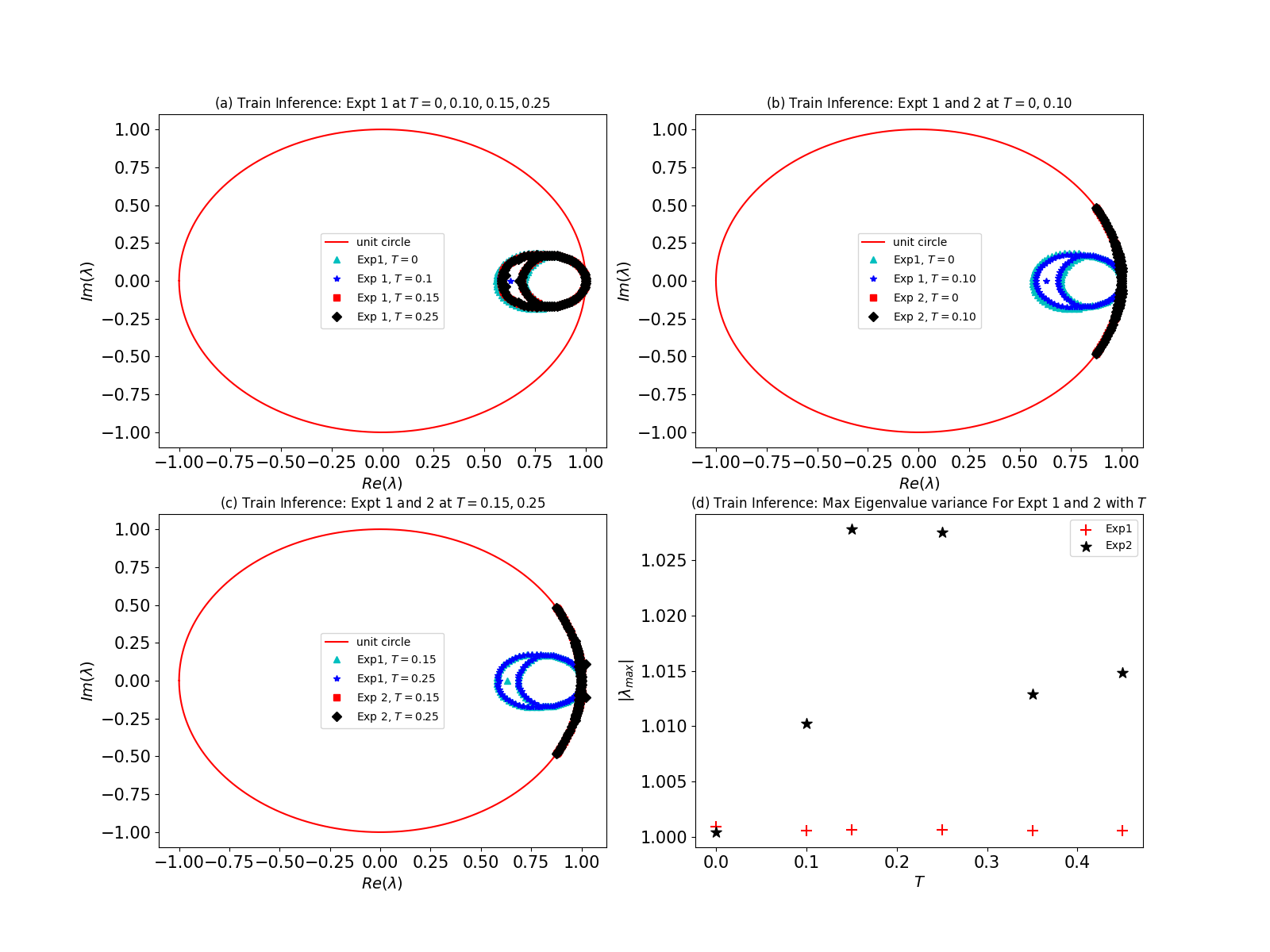}
    \caption{\textbf{Eigenvalues of Burgers Autoregressive model Jacobian in Training set:} Inference on train $max(\phi_{0}) = 2$ shows intrinsic instability of Expt 2 ($p \neq k$) compared to Expt 1 ($p=k$) (a) Eigenvalues of Expt 1 Jacobians at rollouts up to $T=0.25$ show stability, (b) Short-term rollouts at  $T = 0,0.1$ shows Expt 2 with eigenvalues on the unit circle, and less stable than Expt 1 (c) Expt 2 instability at rollouts of $T = 0.15,0.25$ as eigenvalues breach unit circle (d) Rapid instability of Expt 2 eigenvalues at full period rollouts up to $T = 0.5$, while Expt 1 remains bounded.}
    \label{fig:eigen_burgers_train}
\end{figure}

We present the eigenanalysis of the Jacobian in the autoregressive model of the Burgers equation from Section~\ref{sec:results:burgers} for both Expt 1 and 2 when performing inference for train and test $\phi_{0}$. Figure~\ref{fig:eigen_burgers_train} shows the eigenvalues and their position relative to the unit circle when the model performs inference on training $\phi_{0}$. Since this is an autoregressive model, the analysis is performed for the Jacobian at various temporal points in the time window $T=0.5$, corresponding to $n = T/dt = 0.5/0.01 = 50$ timesteps. In Fig.~\ref{fig:eigen_burgers_train}(a), we see that Jacobians of the Expt 1 model are stable even when rolled out to $T=0.25$, as demonstrated by their eigenvalues placed well inside the unit circle. In contrast, in (b), the difference in eigenvalues between the two models when rolled out for $T=0,0.10$ is immediate, as the Expt 2 model lies \textit{on} the unit circle, indicating incipient instability. The impact of this is evident in (c), where the rollout is increased to $T=0.15,0.25$. For the first time, we observe the Expt 2 eigenvalues go outside the unit circle, indicating a clear instability in the model. Finally, we plot the maximum eigenvalue for both models from $T = 0 \rightarrow 0.5$ in (d), where a clear picture emerges: Expt 2 becomes a significantly unstable model as rollout $T$ increases, compared to Expt 1, which largely stays neutral as $|\lambda_{max}| \approx 1.0$ even for the solution at final time $T=0.5$. Expt 2 suffered a significant instability event at $T \approx 0.2$, which continued to $T=0.5$, leading to the predictions seen in Fig.~\ref{fig:pred_burgers_autoreg_train} with large unphysical oscillations. We emphasize that the analysis in this figure was for inference on the training $\phi_{0}$, where the TS truncation term $\frac{\partial^i \phi_{0}^{train}}{\partial x^i}$ is \textit{unchanged in both training and inference}. 

Therefore, the only remaining source of error is from the numerical discretization TS truncation terms is $\sum_{i=q}^{\infty} \frac{\delta x^i}{i!} \frac{\partial^i \phi}{\partial x^i} $ associated with the choice of numerical discretization, which varies for Expt 1 and 2. The results in Section~\ref{sec:results:burgers} and Fig.~\ref{fig:eigen_burgers_train} provide strong evidence to our hypothesis that even in idealized cases where training and test conditions are the same, the models learned are drastically different when the numerical truncation errors between the NeuralPDE and ground truth PDE solver are different.

\begin{figure}[h!]
    \centering
    \includegraphics[width=1.2\linewidth]{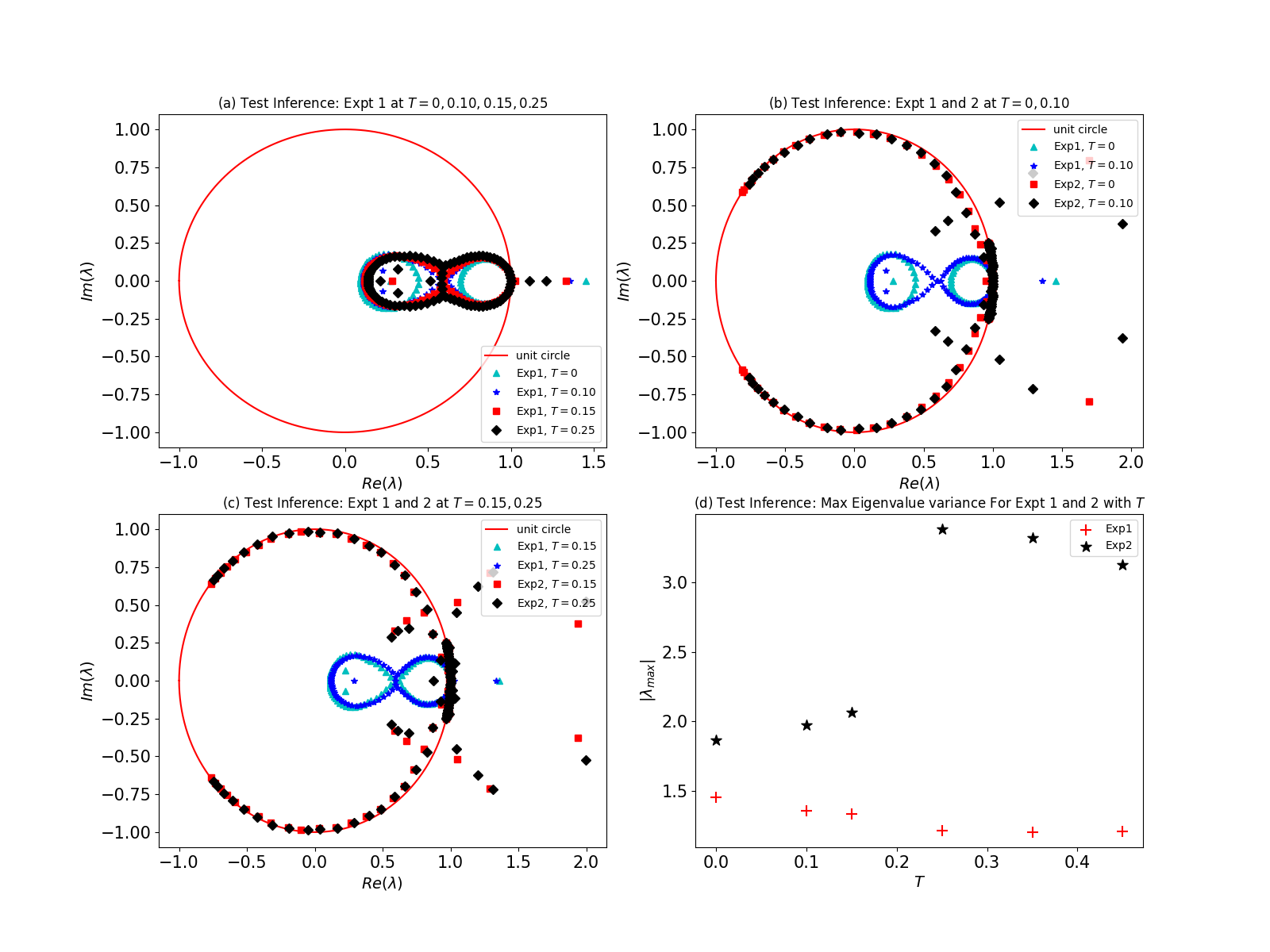}
    \caption{\textbf{Eigenvalues of Burgers Autoregressive model Jacobian in Test set:} Inference on train $max(\phi_{0}) = 10$ shows extreme instability in Expt 2 ($p \neq k$) compared to Expt 1 ($p=k$) (a) Eigenvalues of Expt 1 Jacobians at rollouts up to $T=0.25$ shows some values outside the circle, with majority stable inside, (b) Short-term rollouts at  $T = 0,0.1$ shows Expt 2 with eigenvalues much further from the unit circle than those of Expt 1. Most of Expt 2 eigenvalues are on, or outside, the unit circle, compared to Expt 1, where most are firmly inside. (c) Expt 2 instability at rollouts of $T = 0.15,0.25$ shows vast majority of Expt 2 eigenvalues are more unstable than Expt 1 (d) Extreme instability of Expt 2 eigenvalues at full period rollouts up to $T = 0.5$ compared to Fig.~\ref{fig:eigen_burgers_train}, while Expt 1 shows only weak instability.}
    \label{fig:eigen_burgers_test}
\end{figure}

With this insight, we perform the same study on a case where inference is performed on unseen conditions, and the stability plots are shown in Fig.~\ref{fig:eigen_burgers_test}. Broadly, both Expt 1 and 2 models exhibit instability when the initial condition is far from training, as $\phi^{test}_{0} = 10$. From our mathematical analysis, this is not surprising, since unlike Fig.~\ref{fig:eigen_burgers_train}, there are now two sources of error from both $\frac{\partial^i \phi_{0}^{train}}{\partial x^i}$ and $\sum_{i=q}^{\infty} \frac{\delta x^i}{i!} \frac{\partial^i \phi}{\partial x^i}$.

We see this for Expt 1 in (a), where one eigenvalue is outside the unit circle up to $T = 0.25$. The rest of the eigenvalues have a different distribution than~\ref{fig:eigen_burgers_train}(a), but still inside the unit circle. The stability in (b) paints a dramatically different picture: Both models are stable with eigenvalues far outside the circle, but the Expt 2 eigenvalues are the furthest, and most of its other eigenvalues lie on the perimeter of the unit circle, indicating strong instability even at early time rollouts $T=0,0.1$. By comparison, Expt 1 has roughly two unstable eigenvalues for both $T=0,0.10$. The stability at longer time rollouts up to $T=0.25$ shows similar features, with Expt 2 undergoing a massive instability compared to Expt 1. Finally, the variance of $|\lambda_{max}|$ in (d) shows that both models are unstable with $|\lambda_{max}| > 1.0$ when the initial condition is different from training. However, as the time rollouts increase, we again see the same phenomena as Fig.~\ref{fig:eigen_burgers_train}(d), with Expt 2 undergoing an extreme instability, with the $|\lambda_{max}|_{Expt2}$ more than twice  $|\lambda_{max}|_{Expt1}$. The results in Fig.~\ref{fig:pred_burgers_autoreg_u0_10} demonstrate the effect of this structural instability. While Expt 1 struggles to capture the position of the discontinuity, it captures the rest of the profile satisfactorily. By contrast, Expt 2 shows large unphysical oscillations everywhere in the domain. Expt 2 model completely blows up when $max(\phi_{0}) = 13$, while Expt 1 largely captures the features. Therefore, the two-pronged instability from the initial condition and numerical scheme truncation errors progressively worsen as the gap between the test and train $\phi_{0}$ widens, as evidenced by Fig.~\ref{fig:errorgrowth_burgers_autoreg}, which shows a monotonic increase in prediction error.

\begin{figure}
    \centering
    \includegraphics[width=0.9\linewidth]{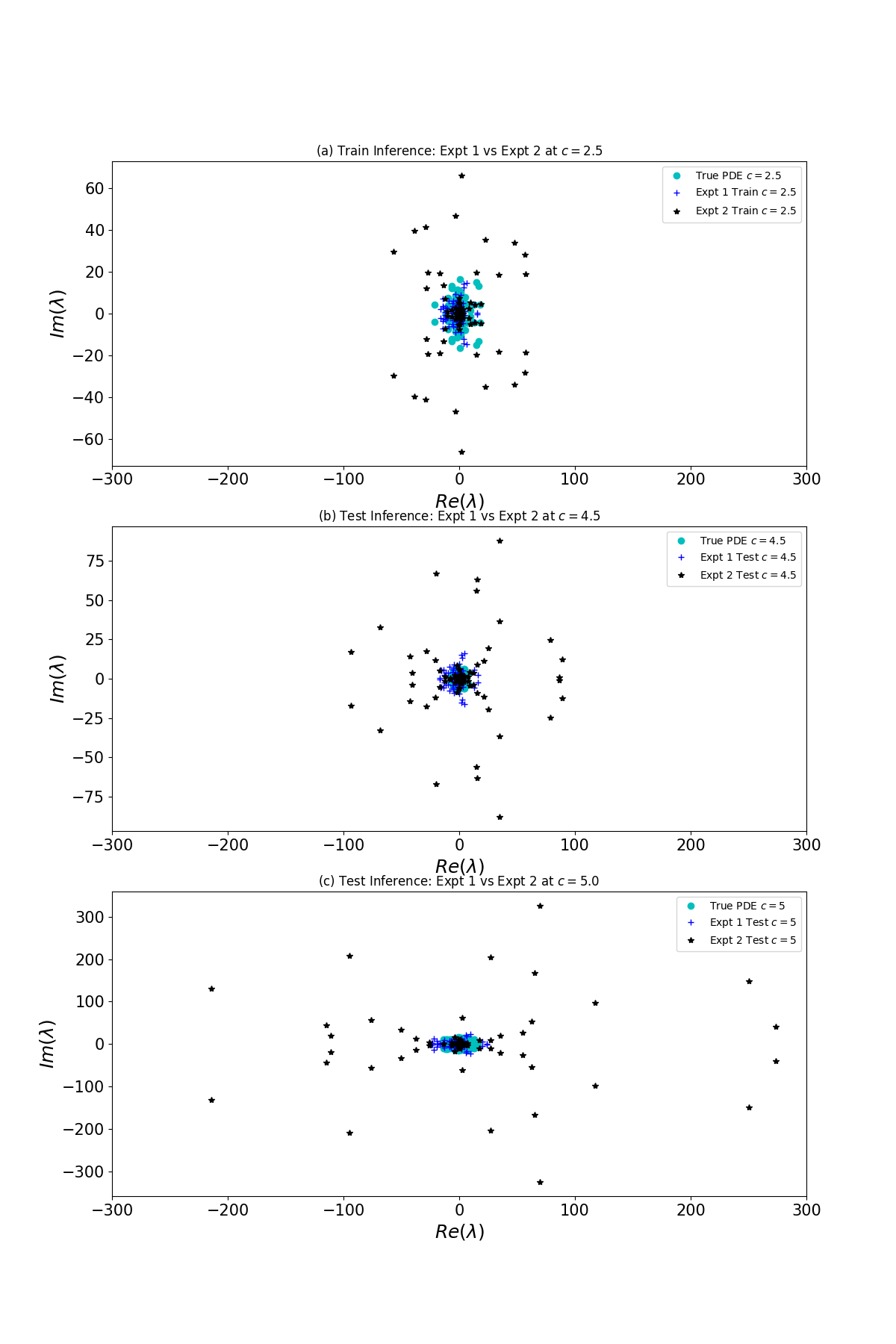}
    \caption{Eigenvalue Stability for the geophysical KdV Equation: Models in Expt 1 and 2 for Train and Testing $c$. In all cases, expt 2 shows maximum eigenvalues three times larger than Expt 1, indicating large instability propagating from $T=0$ to $T=1.0$}
    \label{fig:eigen_gKdV_train_test}
\end{figure}

\begin{figure}[t]
    \centering
    \includegraphics[width=0.75\linewidth]{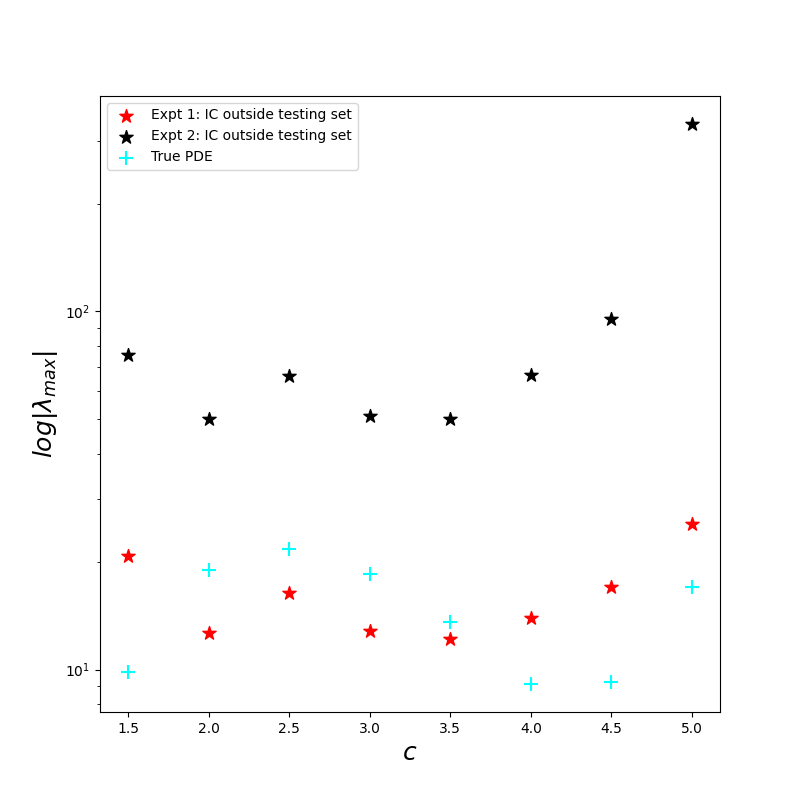}
    \caption{Maximum Eigenvalue $log|\lambda_{max}|$ of the Jacobian $\mathbf{J}_c=\frac{\partial \mathbf{H}\left(\hat{\phi}(t+100\Delta t)\right)}{\partial \hat{\phi(t)}}\bigg \rvert_{t=t_0}$ for Models in Expt 1 and 2 compared with the true Jacobian. Expt 2 model shows log max eigenvalues an order of magnitude larger than Expt 1. Expt 1 tracks the true eigenvalues more closely, albeit with discrepancies.}
    \label{fig:maxeigen_gkdv}
\end{figure}

To demonstrate further evidence for our hypothesis, we perform the same analysis for the models of the gKdV in Section~\ref{sec:results:gkdv}. In this case, the models in Expt 1 and 2 perform one-shot prediction, as opposed to autoregressive in Burgers. One-shot prediction is a popular learning framework in the SciML community as it circumvents some stability challenges in training autoregressive models. However, we demonstrate that it is mathematically susceptible to the same errors. Since this is a one-shot model, we only have access to the cumulative Jacobian $\mathbf{J}_c=\frac{\partial \mathbf{H}\left(\hat{\phi}(t+n\Delta t)\right)}{\partial \hat{\phi(t)}}\bigg \rvert_{t=t_0}$, a stability analysis based on the unit circle is not possible here, as the accumulating rollout for $n = T/ \Delta t = 1.0/0.01 = 100$ timesteps will lead to eigenvalues $>> 1$. However, it still allows us to compare relative magnitudes of $\lambda$ for Expt 1 and 2, as larger values indicate more significant error growth, our metric of interest for model accuracy.

We show the analysis for both experiments at the training $c^{train} = 2.5$ and testing $c^{test} = 4.5, 5.0$ in Fig.~\ref{fig:eigen_gKdV_train_test}. In (a) it is observed that even for training $c$, the eigenvalues for Expt 2 are much more spread out than Expt 1, with $|\lambda_{max}|_{Expt 2} \approx 60$ three times that of Expt 1 $|\lambda_{max}|_{Expt 1} \approx 20$. Expt 1 eigenvalues are closer to the $|\lambda_{max}|$ of the true PDE, which explains its near-perfect prediction in Fig.~\ref{fig:gKdV_pred_train}. For both models, The error growth is exacerbated when the gap between $c^{test}$ and $c^{train}$ increases, as seen in (b), where $c=4.5$. Expt 1 starts diverging from the true eigenvalue distribution, but is eclipsed by Expt 2, where $|\lambda_{max}|_{Expt 2}$ rises to $ \approx 100$ compared to $|\lambda_{max}|_{Expt 1} \approx 25$. In (c), where $c^{test}$ increases by $0.5$ to $c=5.0$ $|\lambda_{max}|_{Expt 2} \approx 300$, indicating that even minor changes in testing initial conditions can have an outsized impact on error growth. An increase for $|\lambda_{max}|_{Expt 1}$ is also observed, but it's limited compared to Expt 2. A complete picture of this error growth is presented in Fig.~\ref{fig:maxeigen_gkdv}, where we plot the variance $log|(\lambda_{max})|_{Expt 1}$ and  $log(|\lambda_{max}|)_{Expt 2}$ with $c$. For both experiments, the least values are seen when $c=2.5$, which is the training condition. For all other $c$, we see higher errors, with $log(|\lambda_{max}|)_{Expt 2}$ almost an order of magnitude higher than $log(|\lambda_{max}|)_{Expt 1}$, with a pronounced error when $c > 4.0$. In summary, we highlight that the eigenanalysis of the gKdV shows remarkable qualitative similarity to the Burgers analysis despite the equations and models being very different because the structural error sources are identical.

\section{Discussion}\label{sec:discussion}
Conventional wisdom and discourse often cast the inherent ``generalization error" in a $NN$ as a potential cause when scientific ML models for PDEs struggle for conditions outside the training set. This work presents the idea that intrinsic errors present in the ``ground truth" from simulations can make or break a model much before we approach generalization error. Importantly, these errors are structural and can be mathematically represented even before a NeuralPDE model is trained. The models in this effort were trained on a trajectory from a single IC, as having more than one obfuscates the impact of a single IC on the TS truncation error. This is often the case when developing models from expensive simulations of complex physics, where only a few trajectories are available (such as climate models). We also show that while both autoregressive and one-shot models may perform differently for the same training datasets, they both suffer from TS truncation errors, as these are \textit{structural errors} inherent in the training data and NeuralPDE discretizations. 

An important contribution is our analysis methodology in studying model accuracy and stability \textit{a priori} using eigen analysis. The Jacobians of the trained models with respect to the initial conditions $\mathbf{J}=\frac{\partial \mathbf{H}\left(\hat{\phi(t)}\right)}{\partial \hat{\phi(t)}}\bigg \rvert_{t=t_0}$ are a valuable \textit{a priori} diagnostic tool that can indicate a NeuralPDE's ability to generalize, and the presence of instabilities that may hinder it from being a robust model. For autoregressive models, this can be obtained at intermediate timesteps, while it can be obtained only at the final timestep for one-shot models. In both these cases, analyzing the eigenvalues of these Jacobians with classical dynamical systems theory can yield insights not possible by only analyzing aggregate error metrics in training. Future efforts will focus on extending this analysis to more complex models.

An assumption we have made often unseen in complex and realistic problems is constant grid size. This simplification was necessary to focus our attention on TS truncation error solely on numerics rather than grid properties. However, numerical simulations of realistic applications and multiphysics problems often have non-uniform grids with different biases, also manifested in the TS truncation error. Typically, a larger grid cell has increased numerical diffusion. Indeed, in wave-like problems different numerical grid sizes can introduce ``impedance matching'' and wave refraction entirely analogous to impedance in electronic circuits \cite{Zlochower12}. To include this grid size variance in the $\varepsilon$ calculation, we must account for the interplay between numerical diffusion between grid size and numerical scheme, and their ``net response", which can be artifacts such as excessive diffusivity or unphysical oscillations from artificial dispersion. Furthermore, we have not focused on errors from time integration due to numerous past efforts in that direction~\cite{brandstetter2022message,lippe2024pde}, and instead keep it constant across models to remove variability, and thereby aiding comparison.

Finally, our effort is restricted to differentiable programming models where $NN$ is part of a model with numerical schemes. Yet, regardless of the type of ML model, our central statement in this work still holds:  The ground truth from PDE simulations is associated with a TS truncation error tied to the numerical discretization and initial conditions, which an ML model implicitly learns. We must now pose a larger question. Can we mathematically represent the error from discretized ground truth on any ML model before training? A rigorous study quantifying the errors incurred by this in the learning process would yield valuable insight into the inherent limits to how much a model can generalize, and aid in uncertainty quantification and trustworthiness. This analysis is complicated because, as seen from this work, it does not have a generic analytic form and is instead dependent on the equation, discretization, and the model. The problem is further compounded when we consider SciML foundation models, which employ large datasets from various numerical solvers.  Our analysis shows the blanket assumption of ``ground truth" extended to this diverse dataset - without seriously considering the mathematical idiosyncracies of the solvers that generated them - can negatively impact ML models. A parallel exists in the computer vision community, where images can be contaminated by a tiny number of pixels that can cause a model to misbehave without explanation. This idea is widely known as an \textit{adversarial attack}~\cite{kurakin2016adversarial,akhtar2018threat}, where datasets with ``invisible" (at least to human eyes) and hard-to-find numerical artifacts  poison~\cite{chen2017targeted,goldblum2022dataset} high-quality datasets to destroy model performance. 

Our investigation unfortunately raises the concern that adversarial attacks (often unintentional) can also  be a factor in large PDE datasets generated by numerical schemes that are unknown to the SciML developer, as is often the case in foundation model research. We have shown that even for a problem as simple as the Burgers equation, the solution generated by two solvers with different finite difference schemes are identical - but the ``invisible" Taylor Series truncated terms can make the model learn different quantities and undergo catastrophic failure. Further still, these datasets can come from various families of numerics: finite difference, spectral, finite element, each with their own numerical approximations, making the error analysis much more complex than that presented in this work. The computational physics community has long considered these factors when designing numerical solvers for verification and validation. There is a long history of numerical analysis that studies the errors of a numerical method. For example, Richardson extrapolation \cite{Richardson1911, Richardson1927}, developed for engineering models of bridges, can be thought of as one of the earliest attempts at addressing this gap. The approach estimates both the form of the true solution (``ground truth'') and the truncation error. This idea may perhaps enable more sophisticated study in NeuralPDEs. If we are to build robust and reliable NeuralPDE models for science, we must extend these models the same rigor~\cite{mcgreivy2024weak} and analysis, and this work is a modest step in that direction.


\backmatter

\bmhead{Acknowledgements}
ATM was supported by the Laboratory Directed Research and Development (LDRD) Early Career Research (ECR) Award and the Artimis Project at Los Alamos National Laboratory(LANL). JMM was supported by LDRD ECR Award number 20220564ECR and the M$^3$AP and AI4C projects at LANL. LANL is operated by Triad National Security, LLC, for the National Nuclear Security Administration of U.S. Department of Energy (Contract No. 89233218CNA000001).
AC was supported by the National Science Foundation (grant no. 2425667) and computational support from NSF ACCESS MTH240019 and NCAR CISL UCSC0008, and UCSC0009. This work is released under Los Alamos National Lab unlimited release policy with document number LA-UR-24-32422.

\begin{appendices}

\section{Methods}\label{app:methods}
This section describes how the training dataset was generated. In both the Burgers and gKdV equations, the terms are discretized using classical finite differences for all the spatial derivatives. All the calculations and training are accomplished with the Julia language, using the Flux~\cite{Flux.jl-2018,innes:2018} machine learning library. Unless mentioned otherwise, we have ensured that the NeuralPDEs and ground truth PDEs are solved exactly the same, to isolate effects of numerical schemes. 

\subsection{Burgers Equation}
For the Burgers equation, the domain has length $L = 2\pi$ with a total evolution time $T = 0.5$. The spatial and temporal resolution are $\delta x = 2\pi / 100$ and $\delta t = 0.01$ respectively. We set the viscosity $\nu = 0.03$ and use $N=100$ grid points throughout this work. The initial condition for this problem is a step function with constant value $max(\phi_{0}) = 2.0$ for the first half of the domain $0 \leq x \leq L/2$ and zero elsewhere. In the test conditions, $max(\phi_{0})$ is varied in steps of 1 from $max(\phi_{0}) = 3.0 \rightarrow 15$.

The neural network model is implemented as a fully connected feed-forward network. It consists of three hidden layers, each with 20 neurons and tanh activation functions, and a final layer mapping to the output grid size of $N=100$. The network is trained to approximate the evolution of the solution over short time windows (auto-regressive learning), with its output integrated into the governing equation.
The true equation is solved numerically to produce training data, while the learned dynamics are used to solve the reduced-order problem. The loss function minimizes the mean squared error (MSE) between the predicted and true solutions at each timestep, with gradients computed using backpropagation via Julia Zygote~\cite{innes2018don}. The network is trained auto-regressively over 3 timesteps to avoid compounding errors, even though inference is performed over the entire simulation duration, corresponding to 50 timesteps. We employ the RK4 integrator using the DifferentialEquations.jl package~\cite{christopher_rackauckas_2024_14043824} in Julia. The model is trained for 1000 epochs and is optimized using the Adam optimizer with a learning rate of $0.01$. Checkpoints save model states, gradients, and optimizer parameters every 20 epochs for computation of Jacobians for subsequent analysis.

\subsection{Geophysical KdV}
The simulations of the geophysical Korteweg-de Vries (gKdV) equation~\cite{geyer2018shallow} were conducted using a spatial domain of length \( L = 5\pi \) discretized into \( N = 256 \) grid cells, resulting in a spatial resolution \( \Delta x = \frac{L}{N} \). Temporal evolution was computed over a total time \( T = 1.0 \) using a time step \( \Delta t = 0.01 \), yielding \( n_t = \frac{T}{\Delta t} \) time steps. The spatial grid points were defined as \( x \in [0, L-\Delta x] \), and the temporal grid spanned \( t \in [0, T] \). Initial conditions were based on a soliton with amplitude \( A = 5.0 \) and velocity \( c = 2.5 \), while a Coriolis factor of \( \omega_0 = 0.5 \) was adopted based on Geyer and Quirchmayr~\cite{geyer2018shallow}. The time integration of the governing equations was carried out using these parameters within the specified time span \( t \in (0, T) \).

The neural network model is a fully connected feed-forward network comprising three hidden layers with $20$ neurons each and tanh activation functions. The model is designed to predict the system's state at the next step while respecting the gKdV equation by incorporating a reduced-order approximation of the third spatial derivative $\partial_x^3$. The training objective minimizes the mean squared error (MSE) between the predicted and true solutions at the final time step of the simulation. Gradients of the loss function are computed using backpropagation with Zygote, and the parameters are updated using the Adam optimizer with a learning rate of $0.01$. Training is conducted over $10000$ epochs for all models. The ground truth solution is generated using high-order central differencing for spatial derivatives and a high-accuracy time integration scheme (Tsit5), and the same scheme is used for model training.

\end{appendices}


\bibliography{sn-bibliography}

\end{document}